\documentclass[conference]{IEEEtran}
\IEEEoverridecommandlockouts
\usepackage{cite}
\usepackage{amsmath,amssymb,amsfonts}
\usepackage{algorithmic}
\usepackage[linesnumbered,ruled,vlined]{algorithm2e}
\usepackage{graphicx}
\usepackage{textcomp}
\usepackage{xcolor}

\usepackage{multirow}
\usepackage{booktabs}
\usepackage[misc]{ifsym}
\usepackage{ntheorem}
\usepackage{float}
\usepackage{subfigure}
\usepackage{threeparttable}
\usepackage{utfsym}

\graphicspath{{./images/}}

\def\eg{{\em e.g.}}
\def\etal{{\em et al.}}

\newcommand{\myPara}[1]{\vspace{.05in}\noindent\textbf{#1}}

\usepackage{setspace}
\SetKwInput{KwInput}{Input}
\SetKwInput{KwOutput}{Output}

\def\BibTeX{{\rm B\kern-.05em{\sc i\kern-.025em b}\kern-.08em
    T\kern-.1667em\lower.7ex\hbox{E}\kern-.125emX}}

\makeatletter
\def\ps@IEEEtitlepagestyle{
\def\@oddfoot{\mycopyrightnotice}
\def\@evenfoot{}
}
\def\mycopyrightnotice{
{\footnotesize 978-1-6654-7189-3/22/\$31.00~\copyright~2022 IEEE\hfill}
\gdef\mycopyrightnotice{}
}

\begin{document}

\title{Privacy-Preserving Student Learning with Differentially Private Data-Free Distillation
}
% \author{Anonymous Submission \#2}
\author{\IEEEauthorblockN{Bochao Liu\IEEEauthorrefmark{1}\IEEEauthorrefmark{2},
Jianghu Lu\IEEEauthorrefmark{1}\IEEEauthorrefmark{2},
Pengju Wang\IEEEauthorrefmark{1}\IEEEauthorrefmark{2}, 
Junjie Zhang\IEEEauthorrefmark{3}, 
Dan Zeng\IEEEauthorrefmark{3},
Zhenxing Qian\IEEEauthorrefmark{4} and
Shiming Ge\IEEEauthorrefmark{1}\IEEEauthorrefmark{2}}
\IEEEauthorblockA{\IEEEauthorrefmark{1}Institute of Information Engineering, Chinese Academy of Sciences, Beijing 100095, China}
\IEEEauthorblockA{\IEEEauthorrefmark{2}School of Cyber Security, University of Chinese Academy of Sciences, Beijing 100049, China}
\IEEEauthorblockA{\IEEEauthorrefmark{3}School of Communication and Information Engineering, Shanghai University, Shanghai 200444, China}
\IEEEauthorblockA{\IEEEauthorrefmark{4}School of Computer Science, Fudan University, Shanghai 200433, China}% <-this % stops an unwanted space
%\thanks{Manuscript received December 1, 2012; revised August 26, 2015. Corresponding author: M. Shell (email: http://www.michaelshell.org/contact.html).}
}

\maketitle

\begin{abstract}
Deep learning models can achieve  high inference accuracy by extracting rich knowledge from massive well-annotated data, but may pose the risk of data privacy leakage in practical deployment. In this paper, we present an effective teacher-student learning approach to train privacy-preserving deep learning models via differentially private data-free distillation. The main idea is generating synthetic data to learn a student that can mimic the ability of a teacher well-trained on private data. In the approach, a generator is first pretrained in a data-free manner by incorporating the  teacher as a fixed discriminator. With the generator, massive synthetic data can be generated for model training without exposing data privacy. Then, the synthetic data is fed into the teacher to generate private labels. Towards this end, we propose a label differential privacy algorithm termed selective randomized response to protect the label information. Finally, a student is trained on the synthetic data with the supervision of private labels. In this way, both data privacy and label privacy are well protected in a unified framework, leading to privacy-preserving models. Extensive experiments and analysis clearly demonstrate the effectiveness of our approach.

%is fed into the teacher model and student model which used for publishing. Since the output of the teacher model is sensitive, we propose a label differential privacy algorithm selective randomized response to protect the privacy information. Finally, we calculate the distillation loss between the output of the student model and the output of the teacher model after selective randomized response processing to train the privacy-preserving student model. In this way, we protect the privacy in the process of knowledge distillation using synthetic data with the labels from the teacher model, leading to a high-performance and privacy-preserving model. Extensive experiments and analysis prove the effectiveness of our approach.

\end{abstract}


\section{Introduction}%1.5 page, include related work
Deep learning models have proven success in many inference tasks \cite{lecun1998poi,krizhevsky2012nips,he2016cvpr,dosovitskiy2020iclr,celeba} by extracting rich knowledge from massive well-annotated data. However, the deployment of these well-performing models may has risk of data privacy leakage since the training data often contain private information~\cite{papernot2017iclr} and the models may be attacked. For example, the existing works~\cite{fredrikson2015ccs,zhang2020cvpr} have shown that the private information in the training data can be obtained from models even if the parameters are not known. Thus, it is very meaningful to design effective solutions that can \emph{learn privacy-preserving models with small accuracy loss}.

Differential privacy~\cite{dwork2006tcc} is one of the most widely used algorithm of privacy protection and provides privacy measurement standard for data and label. Abadi \etal~\cite{abadi2016ccs} first introduced differential privacy into stochastic gradient descent to train deep learning models, and their proposed DPSGD algorithm achieves good differential privacy but leads to large accuracy degradation. Papernot \etal~\cite{papernot2017iclr} proposed private aggregation of teacher ensembles (PATE) that achieves differential privacy by limiting privacy loss with the number of labels. For only label-sensitive setting, Chaudhuri \etal~\cite{chaudhuri2011aclt} proposed the concept of label differential privacy~(LabelDP). Badih \etal~\cite{ghazi2021nips} then introduced prior probabilities into the randomized response and trained privacy-preserving models in a multi-step manner. Malek \etal~\cite{malek2021nips} applied the \cite{papernot2017iclr} and bayesian inference to the label differential privacy setting. Correspondingly, Yuan \etal~\cite{yuan2021nipsws} applied \cite{abadi2016ccs} to the label differential privacy setting and proposed a method named Protocol. Esfandiari \etal~\cite{esfandiari2022aistats} improved the model performance by adding a clustering operation before the randomized response. We found that it is much easier to perform the differential privacy algorithm on only the labels than on the data at the same time, and the model accuracy will be much higher through the above works. However, a major issue is how to convert differential privacy setting into label differential privacy setting.
% So we consider how to apply LabelDP algorithm for training while protecting the data.

To convert differential privacy setting into label differential privacy setting, a key is learning models with generative data and private labels. Recent data-free knowledge distillation can provide this function. Data-free knowledge distillation is a class of approaches which aims to train a student model with a pre-trained teacher model without access to original training data. It uses the information extracted from the teacher model to synthesize data used in the distillation process. Chen \etal~\cite{chen2019iccv} proposed data-free learning for training the student model by exploiting GAN. It uses the teacher model as a fixed discriminator to train a generator to generate the training data used for distillation. Fang \etal~\cite{fang2022aaai} proposed a FastDFKD that applied the idea of meta-learning to the training process to accelerate the efficiency of data synthesis. We found that a slight modification of the generator training method for such methods can learn only the data distribution information and ignore the data representation information.

%Existing researches have led us to recognize that data-free knowledge distillation may play a critical role in solving the problem of converting differential privacy setting to label differential privacy setting. Data-free knowledge distillation is a class of approaches which aims to train a student model with a pre-trained teacher model without access to original training data. It uses the information extracted from the teacher model to synthesize data used in the distillation process. Chen \etal~\cite{chen2019iccv} proposed data-free learning for training the student model by exploiting GAN. It uses the teacher model as a fixed discriminator to train a generator to generate the training data used for distillation. Fang \etal~\cite{fang2022aaai} proposed a FastDFKD that applied the idea of meta-learning to the training process to accelerate the efficiency of data synthesis. We found that a slight modification of the generator training method for such methods can learn only the data distribution information and ignore the data representation information.

\begin{figure*}[!t]
  \centering
  \includegraphics[width=\linewidth]{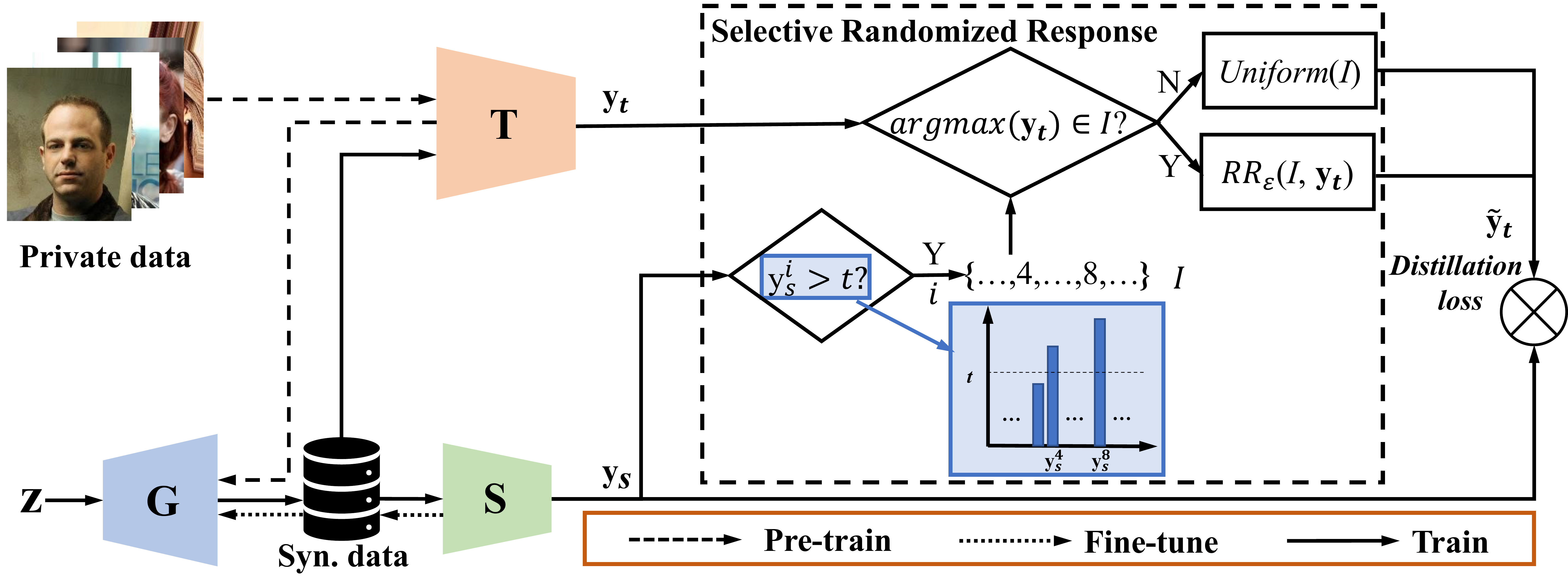}
  \caption{The framework of our differentially private data-free distillation approach. It aims to train a privacy-preserving student model $\textbf{S}$ with teacher-student learning. First, a teacher $\textbf{T}$ is well trained on private data and serves as a fixed discriminator to pre-train a generator $\textbf{G}$ in a data-free manner. Then, massive synthetic data is generated from noisy code $\textbf{z}$ with the generator and fed into the teacher and student $\textbf{S}$ to query differentially private labels with selective randomized response. Finally, with the synthetic data and noisy labels, the student is trained by regressing the teacher knowledge. In this way, both the data privacy and label privacy are well protected in a unified framework, leading to a privacy-preserving student model $\textbf{S}$
  doing the distillation with final labels and outputs of student. In the selective randomized response, we use the output of the student model combined with a threshold $t$ to reduce the number of possible labels and obtain $I$. We implement $\varepsilon$-DP with return $RR_\varepsilon(I, \mathbf{y}_t)$ if correct label in $I$ and $Uniform(I)$ if correct label not in $I$.}
  \label{fig:framework}
\end{figure*}

Inspired by the above works, we propose a privacy-preserving data-free distillation method. As shown in Fig.~\ref{fig:framework}, publishing a model ~(\eg, teacher model) trained directly from private data would compromise privacy, so we treat it as a fixed discriminator to train a generator in a data-free manner. This generator learns only the data distribution to protect the private data. Using this generator implicitly generates data for the distillation process from teacher model to student model. Because querying the teacher model using the generated synthetic data can compromise private information, we propose a LabelDP algorithm selective randomized response to protect the output of the teacher model. The selective randomized response algorithm treats the output of the student model as prior knowledge to reduce the possible output categories to increase the probability of outputting the correct label, and if the possible output does not contain the correct label, a uniform probability distribution is used to reset the possible probability of the output. In summary, our approach can effectively learn privacy-preserving student model by two keys. 
On the one hand, our proposed data-free distillation is able to protect privacy well with the learning of data distribution. The generated synthetic data from this generator will not reveal private information even if it is distributed. On the other hand is that we propose the selective randomized response module to implement DP, which is no longer limited by the number of queries, and introduce the prediction of the student model as prior knowledge for the randomized response. We increase the probability of returning the correct label by setting a threshold, so the student model can learn the knowledge of the teacher model more effectively.

% In summary, our approach can effectively learn privacy-preserving student networks by three key components. First, data-free generator learning is incorporated to generate massive synthetic data. These synthetic data don't need to be protected, and they don't expose the privacy of private data even if recovered by attackers, as we will demonstrate by visualizing them in the experiments section. So we can transform data-sensitive scenario into label-sensitive scenario. Second, our implementation of DP provides strong privacy protection in theory. The label queries are obtained through randomized responses, so privacy can be well protected. Third, we propose the selective randomized response module to implement DP, which is no longer limited by the number of queries, and introduce the prediction of the student model as prior knowledge for the randomized response. We increase the probability of returning the correct label by setting a threshold, so the student model can learn the knowledge of the teacher model more effectively.

Our major contributions are three folds: 1)~we propose a differentially private data-free distillation approach to learn privacy-preserving and high accurate student models via synthetic data, 2) we propose selective randomized response algorithm to privately distill teacher knowledge which provides strong protect label privacy protection in theory, and 3) we conduct extensive experiments and privacy analysis to demonstrate the effectiveness of our approach.

\section{Approach}%1.5 page

\subsection{Problem Formulation}
Given a private dataset $\mathcal{D}$, the goal is to train a student model $\phi_s$ with privacy-preserving capabilities and its accuracy close to the  teacher model $\phi_t$ trained directly on $\mathcal{D}$. To achieve this goal, we propose a privacy-preserving differentially private data-free distillation method. First, we train a teacher model $\phi_t$ directly on $\mathcal{D}$. Then, we use $\phi_t$ as a fixed discriminator to train a generator $\phi_g$ that is used to generate massive synthetic data $\tilde{\mathcal{D}}$. We obtain predictions on $\tilde{\mathcal{D}}$ by querying the teacher model and apply the selective randomized response function which follows $\varepsilon$-LabelDP on them to get labels $\mathcal{L}$. Finally, the student learning can be formulated by minimizing an energy function $\mathbb{E}$:
\begin{equation}
\begin{aligned}
\mathbb{E}(\theta_s; \tilde{\mathcal{D}})&=\mathbb{E}(\phi_s(\theta_s;\tilde{\mathcal{D}}),\mathcal{L})=\mathbb{E}(\phi_s(\theta_s;\tilde{\mathcal{D}}),\mathcal{R}(\phi_t(\theta_t;\tilde{\mathcal{D}}))),
\end{aligned}
\label{eq:problem}
\end{equation}
where $\theta_s$ and $\theta_t$ are the parameters of the student and teacher, respectively. $\mathcal{R}$ is selective randomized response function.
 
From Eq.~\ref{eq:problem}, we can see that our approach can learn privacy-preserving models by two main processes. First, the training of student does not directly access the private data. Second, the labels from the teacher are protected by the selective randomized response module which implements $\varepsilon$-LabelDP. Therefore, privacy leakage can be suppressed very effectively.
During training, the teacher knowledge is transferred to the student through the label $\mathcal{L}$. We solve Eq.~\ref{eq:problem} via two steps, including: 1) data-free generator learning that trains a generator $\phi_g$ with the pre-trained teacher $\phi_t$ as a fixed discriminator to generate synthetic data $\tilde{\mathcal{D}}$, and 2) student learning that applies knowledge distillation is to label the synthetic data with $\phi_t$ and selective randomized response function. And then use these data-label pairs to train the student model $\phi_s$ and fine-tune the generator $\phi_g$. The detailed process is introduced in Alg.~\ref{alg:DP-DFD}.

\begin{algorithm}[ht]
\DontPrintSemicolon
  \caption{Differentially Private Data-Free Distillation (DP-DFD)}\label{alg:DP-DFD}
  \KwInput{
          Number of stages $T$; Pre-trained teacher $\phi_t$; Initial student $\phi_s^{(0)}$; Threshold $t$.
  }
  Training a generator $\phi_g$ with the teacher model $\phi_t$ as a fixed discriminator in data-free manner.\\
  For $i=1$ to $T$:\\
    {\begin{itemize}
        \item[a)] Generate synthetic data $\tilde{\mathcal{D}}^{(i)}$;
        \item[b)] Compute $\mathbf{y}_t$ and $\mathbf{y}_s$ by entering $\tilde{\mathcal{D}}^{(i)}$ into $\phi_t$ and $\phi_s^{(i-1)}$;
        \item[c)] Let $\hat{\mathcal{D}}^{(i)} = [\tilde{\mathcal{D}}^{(i)} , \mathcal{R}(\mathbf{y}_s,\mathbf{y}_t,t)]$;
        \item[d)] Train the student model $\phi_s^{(i)}$ and fine-tune $\phi_g$ on $\hat{\mathcal{D}}^{(i)}$.
    \end{itemize}}
  \textbf{Return} $\phi_s^{(T)}$.\\
  \textbf{Function} selective randomized response $\mathcal{R}$:\\
  \KwInput{Output of student model $\mathbf{y}_s$;
%   =[\mathtt{y}_s^1,\mathtt{y}_s^2,...,\mathtt{y}_s^K]
          The teacher output $y_t$;
          Threshold $t$;
  }
  Select the set $I$ of indexes with condition $y_s^i>t$, and $k=|I|$ is the set size.\\
  \If{$|I|=0$ or $|I|=1$}
    {
        $I$ = [index of top two largest elements in $y_s$]
    }
 \If{$\arg\max(\mathbf{y}_t)\in I$}
    {
        \textbf{Return} $\mathbf{y}_t$ with probability $\frac{e^{\varepsilon}}{e^{\varepsilon}+k-1}$ and the one-hot type of other elements with probability $\frac{1}{e^{\varepsilon}+k-1}$
    }
    \Else
    {
        \textbf{Return} the one-hot type of all elements in $I$ with probability $\frac{1}{k}$
    }
\end{algorithm}

\subsection{Data-Free Generator Learning}
Directly using private data to train the generator will lead to privacy leakage, while using public data will lead to a serious decrease in the accuracy of the student model obtained by distillation, so we want to find a generator training method that does not leak privacy and could match the distribution of the private data. Inspired by~\cite{chen2019iccv}, multi-class classifiers instead of two-class classifiers as discriminators can better learn data distribution, so we adopt a new training approach. We first train a teacher model directly using the private data, and then train a generator using that teacher model as a discriminator with fixed parameters. At the heart of this idea is to take the teacher as a bridge to indirectly learn the distribution of private data. We optimize them by the following loss:
\begin{equation}
\begin{aligned}
\mathcal{L}_g(\tilde{\textbf{x}})=&\ell_{CE}(\phi_t(\theta_t;\tilde{\textbf{x}}), \arg \max\limits_j(\phi_t(\theta_t;\tilde{\textbf{x}}))_j)+\\
&\alpha\phi_t(\theta_t;\tilde{\textbf{x}})\log\phi_t(\theta_t;\tilde{\textbf{x}})+\beta\mathcal{N}(\phi_t,\tilde{\textbf{x}}),
\end{aligned}
\label{eq:generatorloss}
\end{equation}
where $\tilde{\textbf{x}}=\phi_g(\theta_g;\textbf{z})$ generated by $\phi_g$ with parameters $\theta_g$, $\textbf{z}$ is a random vector, $\alpha$ and $\beta$ are the tuning parameters to balance the effect of three terms. The cross entropy $\ell_{CE}(\cdot)$ is used to enforce the outputs of the teacher model closer to the one-hot labels. The smaller it is, the closer the synthetic data distribution is to the private data. The second term is the information entropy loss to measure the class balance of synthetic data. The $\mathcal{N}(\cdot)$ is $l_2$-norm $||*||_2$ to measure the mean and variance of the total synthetic data and the running data fed into the model. In this way, the synthetic data $\tilde{\mathcal{D}}$ generated by the trained generator has a similar distribution to private data without compromising privacy. %We will show some examples in experiments section to demonstrate that no privacy information can be obtained semantically from the synthetic data. 
% Meanwhile, the synthetic data are very helpful for student learning, which can greatly improve accuracy compared to using public data and reduce privacy loss compared to using private data directly.

\subsection{Student Learning with Synthetic Data and Private Labels}

During the training of the student model, we use randomized response~\cite{warner1965asa} for the sensitive labels to achieve DP. $RR_\varepsilon$ mechanism will return correct class label with the probability $\frac{e^{\varepsilon}}{e^{\varepsilon}+K-1}$, and return other labels with probability $\frac{1}{e^{\varepsilon}+K-1}$, where $K$ is the number of classes. To improve the probability of returning the true label without compromising privacy, we introduce the student prediction $\mathbf{y}_s$ and propose selective randomized response algorithm. As shown in function selective randomized response in Alg.~\ref{alg:DP-DFD}, we first set a threshold $t$ and select the set of indexes $I$ with condition $\mathtt{y}_s^i>t$. To ensure the randomness of the output labels, we require that the number of elements in $I$ to be at least 2. We will set $I$ to the set of indexes of top two largest elements in $\mathbf{y}_s$ if the number of elements in $I$ is less than 2. Let $k$ be the number of $I$. If the teacher model's output in $I$, return the $\mathbf{y}_t$ with the probability $\frac{e^{\varepsilon}}{e^{\varepsilon}+k-1}$ and return the one-hot type of other elements with probability $\frac{1}{e^{\varepsilon}+k-1}$~($RR_\varepsilon(I,\mathbf{y}_t)$ in Fig.~\ref{fig:framework}). If the teacher model's output not in $I$, return the one-hot type of the elements in $I$ with probability $\frac{1}{k}$~($Uniform(I)$ in Fig.~\ref{fig:framework}). 

For learning with LabelDP guarantee, we use selective randomized response to randomized outputs from the teacher model for each example of the synthetic data and then apply a general learning algorithm that is robust to random label noise to these data-label pairs. Unlike DPSGD and PATE, which require the composition theorems to calculate the final privacy budget $\varepsilon$, we query the random labels once and reuse them in training process. 
At each stage $i\in [T]$, the synthetic dataset $\tilde{\mathcal{D}}^{(i)}$ is first generated using the generator $\phi_g$ and then enter it into the most recent student model $\phi_s^{(i-1)}$ to obtain $\mathbf{y}_s$ as the prior knowledge. 
We run selective randomized response algorithm with $\mathbf{y}_s$ to obtain the label $\mathcal{L}_i$. We use  $\hat{\mathcal{D}}^{(i)}=\{\tilde{\mathcal{D}}^{(i)},\mathcal{L}_i\}$ to train the student model $\phi_s^{(i)}$ and fine tune the generator $\phi_g$. The loss function for the $i$th epoch is
\begin{equation}
\begin{aligned}
\mathcal{L}_{kd}^{(i)}=\sum\limits_{j=1}\limits^{|\hat{\mathcal{D}}^{(i)}|}\ell_{KL}(\phi_s^{(i)}(\theta_s;\tilde{\textbf{x}}_j),\tilde{\mathbf{y}}_j),~s.t.~(\tilde{\textbf{x}}_j,\tilde{\mathbf{y}}_j)\in \hat{\mathcal{D}}^{(i)},
\end{aligned}
\label{eq:studentloss}
\end{equation}
where $\ell_{KL}(\cdot)$ represents the Kullback-Leibler divergence. For the synthetic dataset $\tilde{\mathcal{D}}=\cup_{i=1}^{T}\tilde{\mathcal{D}}^{(i)}$, iff the size of dataset in each stage is the same, their order will have no effect on the accuracy of student, but $T$ will affect the student accuracy. %We will explore how $T$ affect the student accuracy in experiments.

In our approach, the private data first transfers the knowledge to the teacher model. Directly publishing the teacher model would lead to privacy leakage, so we use the teacher model as a fixed discriminator to train a generator to generate a non-sensitive synthetic dataset with the similar distribution to the private data. We use this dataset to transfer the knowledge from the teacher model to the student model. 
Because only the predictions from the teacher model are sensitive,
we protect the predictions of the teacher model by implementing $\varepsilon$-LabelDP through our propose selective randomized response module. In this way, the knowledge of the private data is transferred to the privacy-preserving student model without access through a data-free distillation approach.

% 2.5 page
\section{Experiments} 
To verify the effectiveness of our differentially private data-free distillation approach~(\textbf{DP-DFD}), we conduct experiments on five datasets and perform comprehensible comparisons with 11 state-of-the-arts. To make the comparisons fair, our experiments use the same settings as these approaches and take the results from their original papers.

\subsection{Experimental Setting}

\vspace{.05in}\noindent\textbf{Datasets.}~The experiments are conducted on five datasets. MNIST~\cite{mnist} and FashionMNIST~(FMNIST)~\cite{fmnist} are both 10-class datasets for $28\times28$ gray handwritten number images and fashion images, respectively. They includes 60K train examples and 10K test examples. CIFAR10 and CIFAR100~\cite{cifar}) consists of 60K $32 \times 32$ color object images in 10 and 100 subjects, where 50K for training and 10K for testing. CelebA~\cite{celeba} contains 202,599 color facial images that are preprocessed by aligning and resizing into $64\times64$. According to hair color and gender attributes, we create two CelebA datasets, CelebA-H and CelebA-G and they uses black/blonde/brown and male/female as labels, respectively. We partition them into training set and test set according to the official criteria~\cite{celeba}. 

\vspace{.05in}\noindent\textbf{Implementation.}~In all experiments, the structure of teacher is $Resnet34$ and we set $\alpha$ and $\beta$ in Eq.~\ref{eq:generatorloss} as 5 and 10 respectively. The structure of student is the same as \cite{wang2021ccs} in data-sensitive experiments and $Resnet18$ in label-sensitive experiments, respectively. For each dataset, we set the threshold value to $1/(2*nc)$ where $nc$ is the number of classes. We evaluate the test accuracy of student under privacy protection.

\subsection{State-of-the-art Comparisons}
\myPara{Comparisons with data-sensitive approaches.}~First, we compare with 7 data-sensitive approaches on MNIST, FMNIST, CelebA-H and CelebA-G under $\varepsilon$=1 and $\varepsilon$=10, including DP-GAN~\cite{xie2018arxiv}, PATE-GAN~\cite{jordon2019iclr}, GS-WGAN~\cite{chen2020nips} and G-PATE~\cite{long2019arxiv}, DP-MERF~\cite{harder2021aistats}, DataLens~\cite{wang2021ccs} and DP-Sinkhorn~\cite{cao2021nips}. 
% We evaluate the test accuracy of student model under the same or lower $\varepsilon$. Here $\varepsilon$ is privacy budget and a smaller budget provides a stronger privacy guarantee.
The results are shown in Tab.~\ref{tab:dp(1_and_10)}. All other approaches are under a failure probability $\delta=10^{-5}$. The accuracy of baseline model which trained directly using private data is $99.21\%$ on MNIST, $91.02\%$ on FMNIST, $93.53\%$ on CelebA-G and $88.68\%$ on CelebA-H. From Tab.~\ref{tab:dp(1_and_10)}, we can see that our DP-DFD shows substantially higher performance than other approaches especially when $\varepsilon=1$. In particular, the accuracy of our DP-DFD outperforms the other best-performing approaches by 21 percentage points. When $\varepsilon=10$, our DP-DFD also has an absolute advantage and is at least 13 percentage points higher than the other approaches. Even for high-dimensional datasets like CelebA-G and CelebA-H, our DP-DFD still shows the state-of-the-art performance, which also demonstrates the advantage of DP-DFD over other privacy-preserving approaches on high-dimensional datasets.

\myPara{Comparisons with label-sensitive approaches.}~Then, we compare with 4 LabelDP approaches (LP-MST~\cite{ghazi2021nips}, ALIBI~\cite{malek2021nips}, ClusterRR~\cite{esfandiari2022aistats} and Protocol~\cite{yuan2021nipsws}) on MNIST, FMNIST, CIFAR10 and CIFAR100 under the same $\varepsilon$. The results are shown in Tab.~\ref{tab:LabelDP}. We can find that our method performs optimally for all four datasets and for different $\varepsilon$. In particular, it achieves a correct rate of 74.67 when $\varepsilon$ equals 8 on the CIFAR100 dataset, surpassing many methods that use the raw data for direct distillation. The effectiveness of our method is further demonstrated by the fact that our method has higher performance than the other four methods trained directly using raw data when there is no restriction on the amount of generated synthetic data.
\begin{table*}[!htbp]
\caption{Accuracy comparisons with 7 data-sensitive approaches: Test accuracy under different privacy budget $\varepsilon$.}\label{tab:dp(1_and_10)}
    \small
 	\setlength{\tabcolsep}{2.8pt}%
 	\renewcommand\arraystretch{1.1}
	\begin{center}
		\begin{threeparttable}
			\begin{tabular}{l|c|c|cccccccc}
				\toprule
				\textbf{Dataset}& \textbf{Baseline} &$\varepsilon$ & \textbf{DP-GAN} & \textbf{PATE-GAN} & \textbf{G-PATE} & \textbf{GS-WGAN} & \textbf{DP-MERF} & \textbf{DataLens} & \textbf{DP-Sinkhorn} & \textbf{DP-DFD}\cr
				\midrule
				\multirow{2}{*}{\textbf{MNIST}} & \multirow{2}{*}{\textbf{0.9921}} & 1 & 0.4036 & 0.4168 & 0.5810 & 0.1432 & 0.6500 & 0.7123 & - & \textbf{0.9762} \cr
				& & 10 & 0.8011 & 0.6667 & 0.8092 & 0.8075 & 0.6870 & 0.8066 & 0.8320 & \textbf{0.9856}\cr
				\midrule
				\multirow{2}{*}{\textbf{FMNIST}} & \multirow{2}{*}{\textbf{0.9102}} & 1 & 0.1053 & 0.4222 & 0.5567 & 0.1661 & 0.6100 & 0.6478 & - & \textbf{0.8917} \cr
				& & 10 & 0.6098 & 0.6218 & 0.6934 & 0.6579 & 0.6250 & 0.7061 & 0.7110 & \textbf{0.9074}\cr
				\midrule
				\multirow{2}{*}{\textbf{CelebA-G}} & \multirow{2}{*}{\textbf{0.9353}} & 1 & 0.5330 & 0.6068 & 0.6702 & 0.5901 & - & 0.7058 & - & \textbf{0.7814} \cr
				& & 10 & 0.5211 & 0.6535 & 0.6897 & 0.6136 & 0.6500 & 0.7287 & 0.7630 & \textbf{0.8934}\cr
				\midrule
				\multirow{2}{*}{\textbf{CelebA-H}} & \multirow{2}{*}{\textbf{0.8868}} & 1 & 0.3447 & 0.3789 & 0.4985 & 0.4203 & - & 0.6061 & - & \textbf{0.6753} \cr
				& & 10 & 0.3920 & 0.3900 & 0.6217 & 0.5225 & - & 0.6224 & - & \textbf{0.8207}\cr
				\bottomrule
			\end{tabular}
		\end{threeparttable}
	\end{center}
\end{table*}

\begin{table}[!htbp]
\caption{Accuracy comparisons with 4 Label sensitive approaches: Test accuracy under different privacy budget $\varepsilon$.}\label{tab:LabelDP}
    \small
 	\setlength{\tabcolsep}{1.8pt}%
 	\renewcommand\arraystretch{1.1}
	\begin{center}
		\begin{threeparttable}
			\begin{tabular}{l|c|ccccc}
				\toprule
				\textbf{Dataset} &$\varepsilon$ & \textbf{LP-2ST} & \textbf{ALIBI} & \textbf{ClusterRR} & \textbf{Protocol} & \textbf{DP-DFD}\cr
				\midrule
				\textbf{MNIST} & 1 & 0.9582 & - & 0.9000 & - & \textbf{0.9762} \cr
				\midrule
				\textbf{FMNIST} & 1 & 0.8326 & - & 0.8800 & - & \textbf{0.8917} \cr
				\midrule
				\multirow{2}{*}{\textbf{CIFAR10}} & 1 & 0.6367 & 0.8420 & 0.6857 & - & \textbf{0.8796} \cr
				& 2 & 0.8605 & - & - & 0.8184 & \textbf{0.8812}\cr
				\midrule
				\multirow{2}{*}{\textbf{CIFAR100}} & 3 & 0.2874 & 0.5500 & - & - & \textbf{0.5861} \cr
				& 8 & 0.7410 & 0.7440 & - & - & \textbf{0.7467}\cr
				\bottomrule
			\end{tabular}
		\end{threeparttable}
	\end{center}
\end{table}

\subsection{Ablation Studies}
After the promising performance is achieved, we further analyze each influencing factor in our approach, including the impact of loss terms in the data-free generator learning, the amount of synthetic data and the number of stages.

\myPara{Loss function.} To further understand the improvement of each component of the loss function during data-free training of the generator, we designed experiments on MNIST and FMNIST under $\varepsilon$=10 to explore the contribution of each component. The results are shown in Tab.~\ref{tab:loss}. where CE means the cross entropy loss term, IE is the information entropy loss term and Norm is the normalized term for the mean and variance of the data. We can see that the normalization term of the data has the greatest impact, followed by the information entry loss term and finally the cross entropy loss term. We speculate that this may be related to the randomness of the data generated by the generator, which limits the distribution of the data to make the generated synthetic data more usable, so it has a greater impact on the accuracy of the student model.
\begin{table}[!htbp]
\small
\caption{Impact of loss terms in training generator under $\varepsilon$=10.}\label{tab:loss}
 	\setlength{\tabcolsep}{7.6pt}%
 	\renewcommand\arraystretch{1.15}
	\begin{center}
		\begin{threeparttable}
			\begin{tabular}{l|ccc|c}
				\toprule
				\textbf{Dataset}& \textbf{CE} & \textbf{IE} & \textbf{Norm} & \textbf{Accuracy} \cr
				\midrule
				\multirow{4}{*}{\textbf{MNIST}} & \usym{2713} & \usym{2713} & \usym{2713} & \textbf{0.9856} \cr
				\cline{2-5}
				& \usym{2717} & \usym{2713} & \usym{2713} & 0.9655 \cr
				& \usym{2713} & \usym{2717} & \usym{2713} & 0.9432 \cr
				& \usym{2713} & \usym{2713} & \usym{2717} & 0.8801 \cr
				% \midrule
				% \multirow{4}{*}{\textbf{FMNIST}} & \usym{2713} & \usym{2713} & \usym{2713} & \textbf{0.8974} \cr
				% \cline{2-5}
				% & \usym{2717} & \usym{2713} & \usym{2713} & 0.8804 \cr
				% & \usym{2713} & \usym{2717} & \usym{2713} & 0.8794 \cr
				% & \usym{2713} & \usym{2713} & \usym{2717} & 0.8122 \cr
				\bottomrule
			\end{tabular}
		\end{threeparttable}
	\end{center}
\end{table}

\begin{figure}[t]
  \centering
  \includegraphics[width=\linewidth]{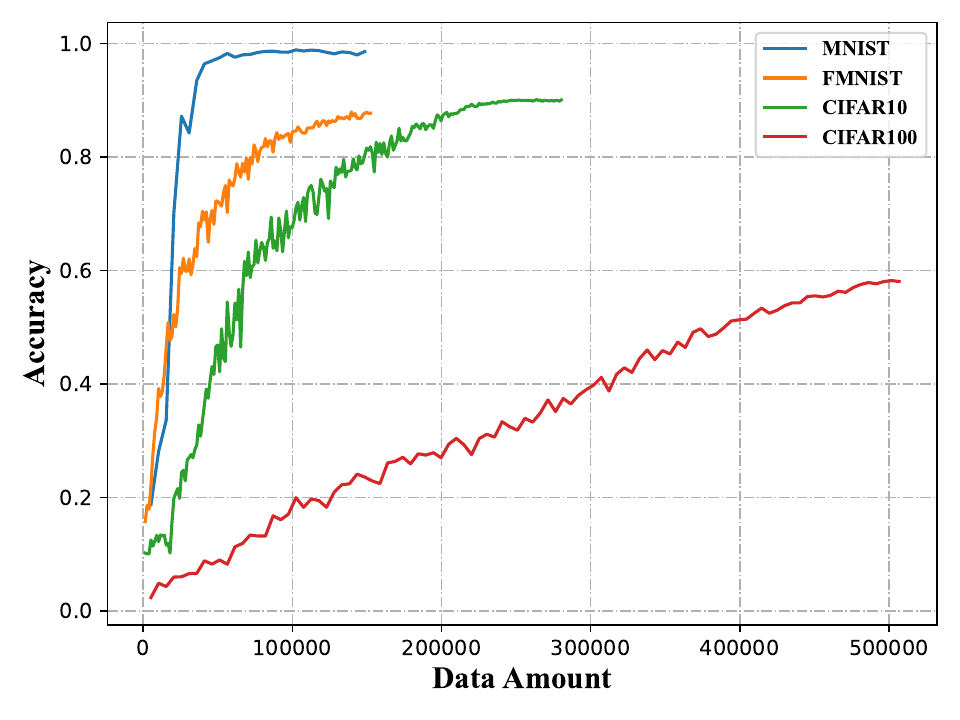}
  \caption{The effect of different amount of synthetic data ($\varepsilon$=1).}
  \label{fig:effect_of_data_amount}
\end{figure}

\myPara{Data amount.} 
We further conducted experiments on MNIST, FMNIST, CIAFR10 and CIFAR100 datasets under $\varepsilon=1$. The results are shown in Fig.~\ref{fig:effect_of_data_amount}. We found that MNIST dataset converges at about 50,000 data volume, FMNIST converges at about 120,000, CIFAR10 and CIFAR100 converge at about 220,000 and 500,000, respectively. As the difficulty of datasets increases, the amount of data required to achieve convergence increases. We suspect that this is because the more difficult the dataset is, the more difficult its distribution knowledge is to learn, so the larger the amount of data required. We note that the CIFAR10 dataset is more difficult than FMNIST, but the reason why CIFAR10's final accuracy is similar to FMNIST's is that the network structure is different.

\begin{figure}[t]
  \centering
  \includegraphics[width=\linewidth]{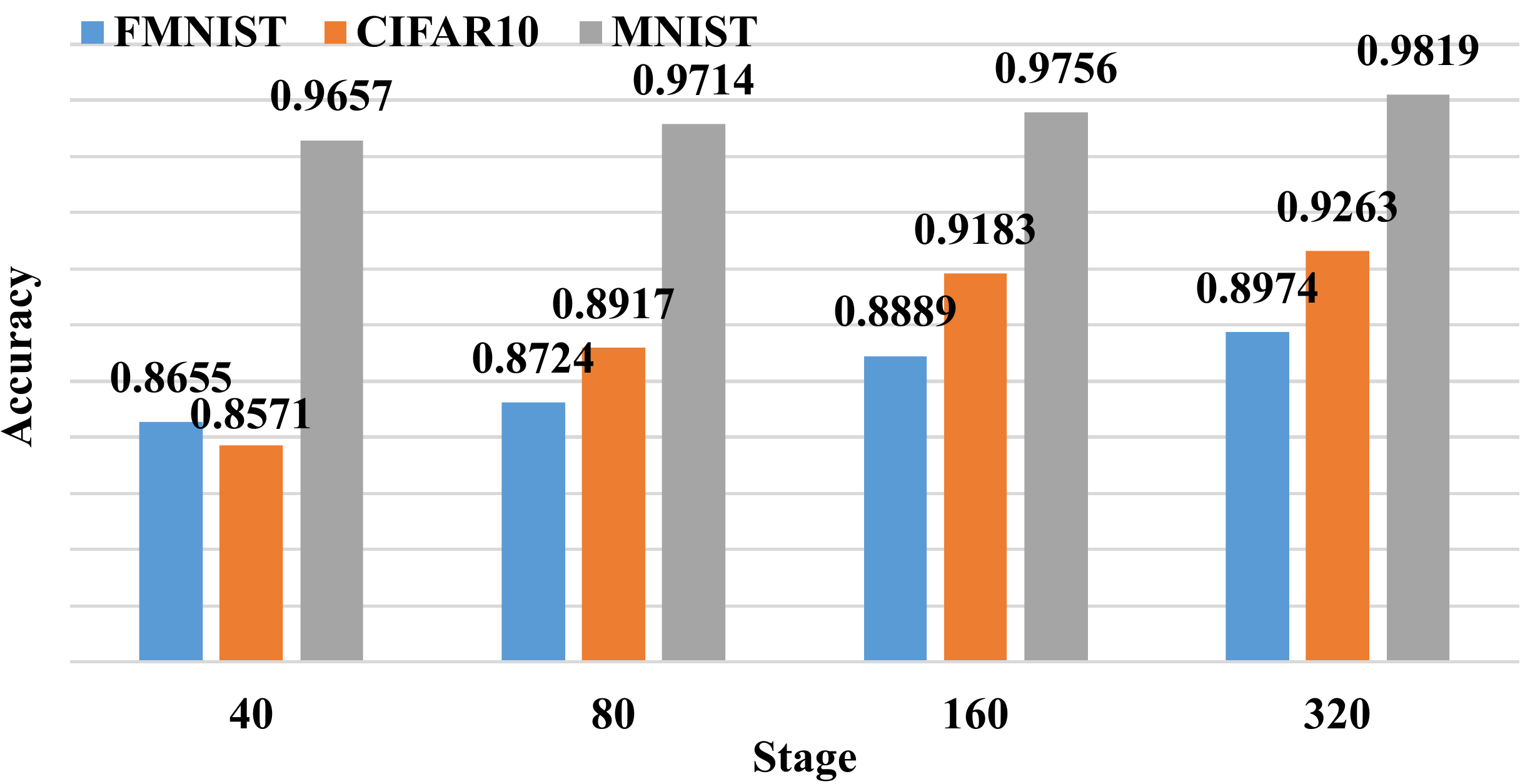}
  \caption{The effect of different number of stages ($\varepsilon$=10).}
  \label{fig:effect_of_stage}
\end{figure}

\myPara{Number of stages.} To explore the effect of the number of stages, we conducted experiments on MNIST, FMNIST and CIFAR10 datasets under $\varepsilon$=10. The results are shown in Fig.~\ref{fig:effect_of_stage}. Experimental results show that between 20 and 320, the accuracy of the student model increases with the increase of stages. As the classification difficulty of MNIST, FMNIST and CIFAR10 datasets increases, the effect of stages becomes greater. The experimental results are as we expected because we used the prediction of the student model as the prior knowledge. As the training process proceeds, the more accurate the prediction of the student model becomes, which means the higher the probability of outputting the correct label. The greater the percentage of synthetic data being correctly labeled, the better the student model performance will be.

\subsection{Privacy-Preserving Analysis}
\begin{figure}[t]
  \centering
  \includegraphics[width=\linewidth]{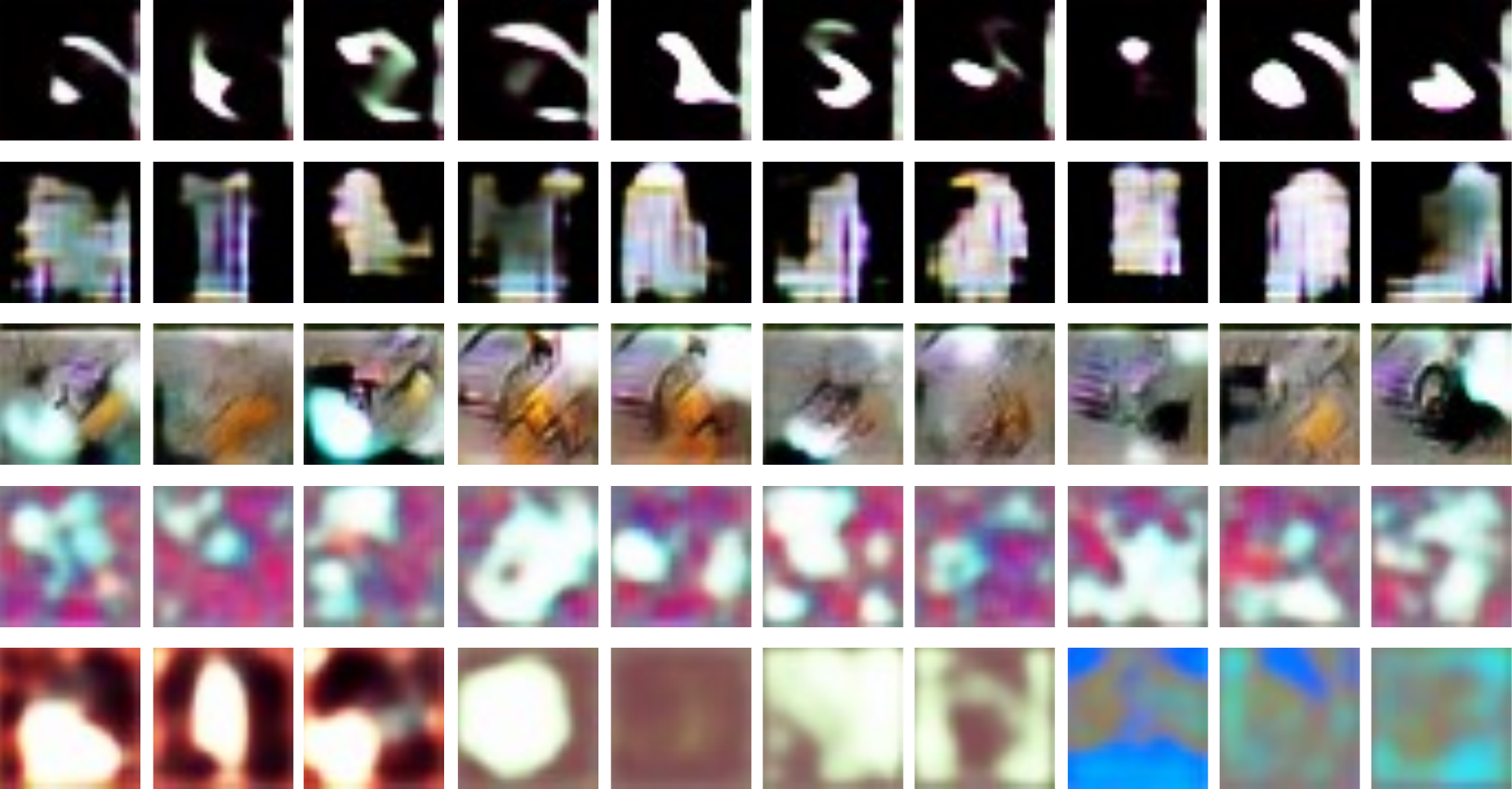}
  \caption{The examples of the generated synthetic data. From top to bottom: MNIST, FMNIST, CIFAR10, CelebA-G and CelebA-H.}
  \label{fig:visualization}
\end{figure}

\vspace{.05in}\noindent\textbf{Data generation.}
To demonstrate that the direct use of synthetic data in our approach doesn't leak information of private data, we visualize some examples for MNIST, FMNIST, CIFAR10 and CelebA, as shown in Fig.~\ref{fig:visualization}. The first row is MNIST, followed by FMNIST, CIFAR10, CelebA-G and CelebA-H in that order. We found that even for the simplest MNIST synthetic data, we could not semantically identify it as a handwritten font. Despite its inability to be recognized by humans, it has high utility in terms of training high performance models. We also found something interesting: such synthetic data can train a model that performs well, which raises an interesting question about what machine learning models actually learn from data?

\begin{figure}[!t]
  \centering
  \includegraphics[width=\linewidth]{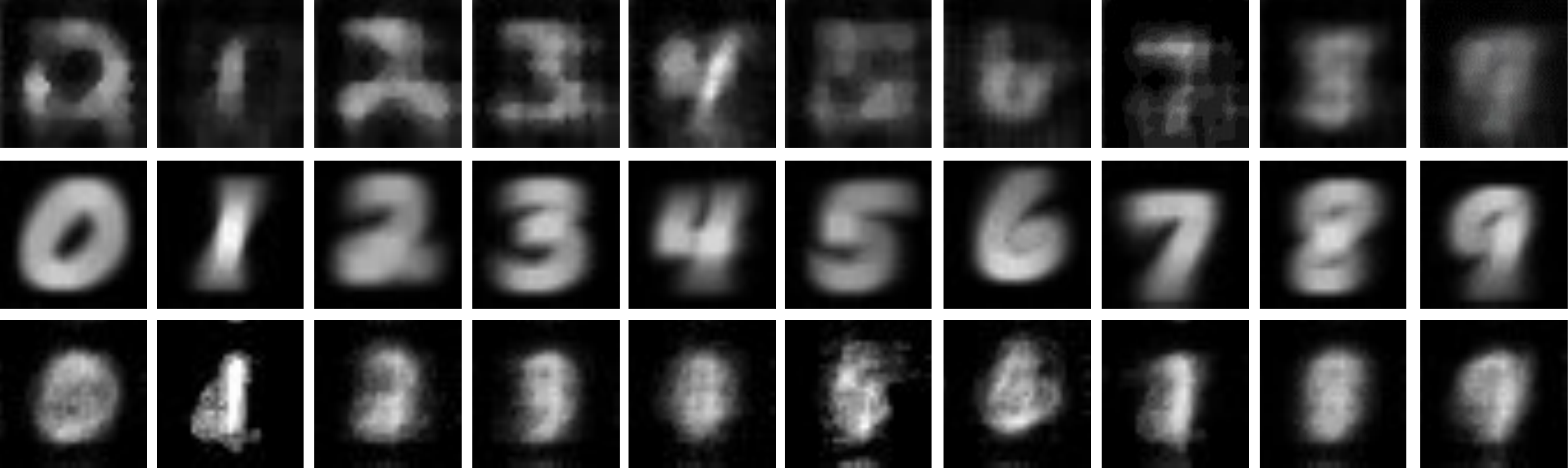}
  \caption{The results of model-inversion attack against the students trained on MNIST with DataLens (top),  ALIBI (middle) and DP-DFD (bottom).}
  \label{fig:attack}
\end{figure}

\vspace{.05in}\noindent\textbf{Model-inversion attack.}
We perform a model-inversion attack~\cite{zhang2020cvpr} on a typical data-sensitive approach and a label-sensitive approach to further demonstrate that our approach can protect data privacy. The results are shown in Fig.~\ref{fig:attack}. The first row is the results of the attack on a typical data-sensitive method DataLens~\cite{wang2021ccs}, while the second row shows the results of the attack on a typical label-sensitive method ALIBI~\cite{malek2021nips}. The last row is the results of the attack on our DP-DFD. We emphasize that the authors of \cite{zhang2020cvpr} stress in their original paper that differential privacy hardly works against this attack method, but we can find that even for experiments on the simplest MNIST dataset, our method still can defend against this attack and protect the privacy of the private data.

\section{Conclusion}
Typically, publishing deep learning models may pose the risk of privacy leakage. To facilitate model deployment, we propose a differentially private data-free distillation approach (DP-DFD) that does not use private data in the training process of publish model. This approach uses the teacher model trained directly with private data as a bridge to transfer knowledge from private data to publish model. The generator trained in a data-free manner can learn the distribution of the private data and enhance the knowledge of the publish model to compensate for the loss of the accuracy without compromising privacy. In addition, we also provide differential privacy analysis for our selective randomized response and DP-DFD to demonstrate that it provides strong privacy guarantees in theory. We have conducted extensive experiments and analyses to show the effectiveness of our approach. In the future, we will explore the approach in more practical applications, such as federated learning on medical images and financial data.

\myPara{Acknowledgements.}~This work was partially supported by grants from the Beijing Natural Science Foundation (19L2040) and National Key Research and Development Plan (2020AAA0140001).  

%\section*{References} % 0.5 page

\end{document}